\documentclass[11pt,a4paper]{article}
\usepackage[hyperref]{emnlp2018}
\usepackage{times}
\usepackage{latexsym}

\usepackage{graphicx}
\graphicspath{ {images/} }

\usepackage{url}

\aclfinalcopy 

\title{Adversarial Text Generation Without Reinforcement Learning}

\author{David Donahue \\
  University of Massachusetts Lowell \\
  {\tt david\_donahue@student.uml.edu} \\\And
  Anna Rumshisky \\
  University of Massachusetts Lowell \\
  {\tt arum@cs.uml.edu} \\}

\date{}

\begin{document}
\maketitle
\begin{abstract}
Generative Adversarial Networks (GANs) have experienced a recent surge in popularity, 
performing competitively in a variety of tasks, especially in computer vision.
However, GAN training has shown limited success in natural language processing. This is largely because sequences of text are discrete, and thus gradients cannot propagate from the discriminator to the generator. Recent solutions use reinforcement learning to propagate approximate gradients to the generator, but this is inefficient to train. We propose to utilize an autoencoder to learn a low-dimensional representation of sentences. A GAN is then trained to generate its own vectors in this space, which decode to realistic utterances. We report both random and interpolated samples from the generator. Visualization of sentence vectors indicate our model correctly learns the latent space of the autoencoder. Both human ratings and BLEU scores show that our model generates realistic text against competitive baselines. 
\end{abstract}

\section{Introduction}



\begin{figure*}[ht]
\begin{center}
\includegraphics[scale=.08]{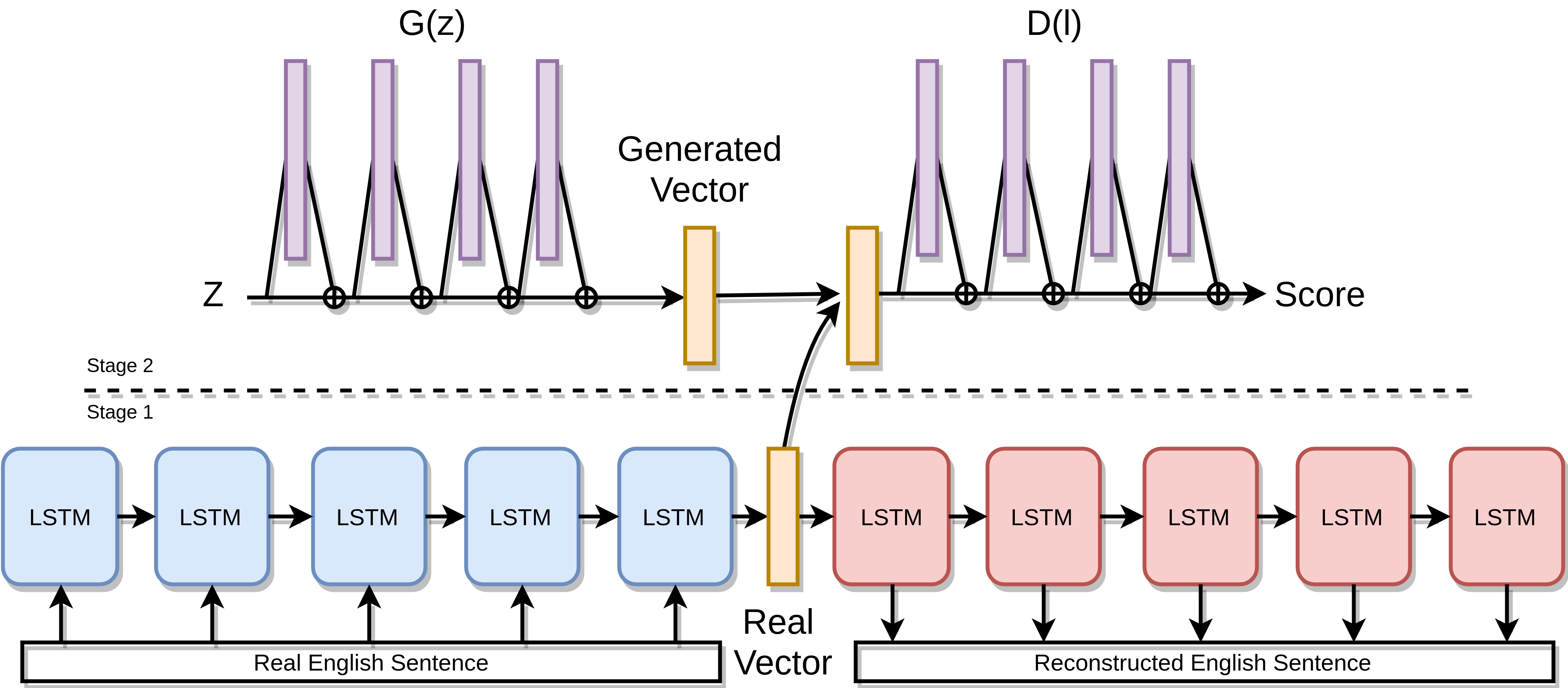}
\end{center}
\caption{LaTextGAN model architecture. The discriminator network $D(l)$ receives sentence representations produced by the fully-trained autoencoder, and from the generator network $G(z)$.}
\label{fig:model}
\end{figure*}



Over the past several years, deep learning models have provided large performance gains on many tasks requiring language generation, from machine translation \cite{johnson2016google} to dialogue agents \cite{serban2016building} to summarization \cite{rush2015neural} and question answering \cite{weissenborn2017making}. A recent 2018 survey paper \cite{gatt2018survey} includes the discussion of neural approaches to natural language generation (NLG) in three out of four chapters dedicated to the current state-of-the-art methods.



The probabilistic neural language model (NLM) is one important example. NLMs have long been utilized for language generation, by predicting sequence probabilities from learned word representations \cite{bengio2001neural}. These models generate text without being conditioned on any input, by outputting a distribution over the vocabulary at each time step which is sampled and given as input at the next time step.
More recently, variational autoencoders (VAEs) have been used for sentence generation \cite{bowman2015generating}. VAE models enforce a prior distribution on the latent output of the encoder, to smooth the space. Random points selected in this space then decode to valid sentences. However, the latent space is not always uniform \cite{makhzani2015adversarial}, and generated examples cannot be conditioned on input features. 

Generative Adversarial Networks were recently proposed as a method for image generation \cite{goodfellow2014generative}. 
GANs have been very successful in computer vision, where they have been applied to a variety of tasks from image captioning \cite{zhang2017stackgan} to image super-resolution \cite{ledig2016photo}. 
Interestingly, while 
GANs have shown exceptional promise for generating realistic data, applying them to text has proved very difficult, largely because text is discrete, and thus gradients cannot propagate from the discriminator to the generator.
%

Developing methods that overcome this obstacle and leverage GANs for text generation is the focus of this paper. Autoregressive models such as the RNN produce a sequence one token at time, by sampling from a generated distribution over the vocabulary at each time step. This sampling occurs at the final layer of the model. However, introducing variation at the final layer can hinder higher-level sentence planning \cite{serban2017hierarchical}. In contrast, GANs insert variation starting from the input layer, which encourages the model to generate in a top-down manner. This is one motivation for the long-term application of GANs to NLG. In addition, Adversarial training of a GAN happens at the sequence level, instead of at the word level. This may encourage greater textual coherence.

It is not currently possible to train a GAN to produce text directly using standard optimization, as text is discrete and thus gradients cannot be passed from discriminator to generator during training. To overcome optimization difficulties involving the discrete text output of the generator, \newcite{yu2017seqgan} utilize reinforcement learning by directly applying policy gradients to the generator. \newcite{zhang2017adversarial} use a soft-argmax approximation with a convolutional network discriminator to smooth policy gradients to the generator, and pre-train the discriminator on sentence permutations to speed-up convergence. \newcite{li2017adversarial} train a discriminator to score only partially decoded sequences. 
To reduce the instability of sequence prediction with recurrent neural networks, \newcite{lamb2016professor} use an adversarial network to encourage similar system behavior during training and prediction. In that work, the adversarial model acts as a training regularizer for a RNN decoder, but is not used during prediction.

Given recent developments in the literature, we propose a Generative Adversarial Network model for sentence generation which does not require reinforcement learning. To overcome the discrete nature of text, we utilize an autoencoder (AE) to encode sentences into smooth sentence representations. A generator network is then trained to generate its own sentence representations in the learned latent space. Each sentence vector produced by the generator is then passed through the AE decoder, which decodes to the nearest sentence. We evaluate our system against multiple baselines and show our generated sentences score well on both human and automatic methods.



\section{LaTextGAN for Sequence Generation}

We introduce LaTextGAN (latent-space GAN for text) for the purpose of generating discrete sequences. In this paper, we focus specifically on the construction and application of LaTextGAN to the unconditional generation of English sentences. Figure \ref{fig:model} contains a diagram of our proposed model. We utilize an autoencoder component which learns a dense low-dimensional representation of text. 
A generator network is utilized to produce additional points in this latent variable space, which decode to valid sentences. As is typical for Generative Adversarial Networks, a discriminator network is trained to classify real and generated sentences from their latent representations. The generator attempts to fool the discriminator by generating more realistic sentence representations. 
\subsection{Textual Autoencoder}

Autoencoders are designed to learn a low-dimensional representation of text by using an encoder network to compress information about each sentence into a finite vector. A decoder network is tasked with reconstructing the input representation from the vector. We utilize a Long-Short Term Memory (LSTM) network 
for both the encoder and decoder \cite{hochreiter1997long}. 
The LSTM network reads each sentence sequentially, one word at a time. 

During sentence reconstruction, the decoder takes the encoder latent representation and the previous hidden state as input and produces a probability distribution which is 
used to select the word at that time-step. We use greedy sampling in our autoencoder, and select the highest probability word at each timestep. 

\begin{table*}[ht]
\centering
\small
\begin{tabular}{ |l|l| } 
 \hline
 LaTextGAN Sampled Sentences & LaTextGAN Interpolations in Input Space \\
 \hline
 He lifted his hand. & It was as late when everyone else was waiting.\\
 He arrived at the subject. & They told them we ate all.\\
 The music hit by an instant and released her hair. & They knew how soon these.\\
 You ve always been in the world to get my friends. & They found them then.\\
 The clock pulled his head away. & You keep them then.\\
 Jill pointed at the door and stared at them blankly . & They found her hands.\\
 Egwene finished by the dwarf. & They sat beside her.\\
 He knew she would be able to keep her alive. & They sat on captain.\\
 I thought you would be there. & He looked at her.\\
 The words he could feel. & He watched her.\\
 \hline
\end{tabular}
\caption{Left: LaTextGAN sentences decoded using random Gaussian vectors as input. Right:
LaTextGAN sentences decoded by moving linearly through input space.}
\label{table:examples}
\end{table*}

\begin{table*}[ht]
\centering
\small
\begin{tabular}{ |r|r|r|r|r| } 
 \hline
 Model & More Realistic & Less Realistic & Equally Realistic & BLEU Score\\
 \hline
 LaTextGAN & \bf{13.9}\% & 55.6\% & 30.5\% & 0.678\\
 NLM & 12.2\% & 46.6\% & \bf{41.2}\% & 0.643\\
 VAE & 6.0\% & 81.6\% & 12.4\% & \bf{0.688}\\

 \hline
\end{tabular}

\caption{Human evaluation of model-generated sentences as more realistic, less realistic, or equally realistic with respect to real English sentences. BLEU-4 score calculated on a held-out validation set.}
\label{table:human_eval}
\end{table*}

\subsection{GAN Architecture Overview}

It seems natural to model both the generator and discriminator using standard fully-connected networks. However, randomly-initialized fully-connected layers are notoriously harder to train as layer depth increases. To mitigate gradient instability associated with these networks, we instead represent the generator and discriminator each using a ResNet architecture \cite{he2016deep}.

 


\subsection{Training Procedure}

To aid in training quality, we adopt the Improved Wasserstein GAN network from \newcite{gulrajani2017improved}, a modification of the original Wasserstein GAN \cite{arjovsky2017wasserstein} which formulates the training objective as:

\begin{equation}
\max_{\theta} E_{z~p(z)}[f_w(g_{\theta}(z)] - E_{x~p(x)}[f_w(x)]
\end{equation}

\noindent
for discriminator (critic) $f_w$ and generator $g_{\theta}$. \newcite{gulrajani2017improved} also apply this training objective to large ResNet architectures, reassuring their applicability to this task. 

\section{Evaluation}

\subsection{Toronto Book Corpus}

While large corpora are usually biased toward particular domains, books offer a variety of genres and dialects. 
%
While it is impossible to obtain sentences that exactly match the distribution of the English language (datasets are always biased toward particular domains), books offer a wide variety of sentences from different genres and dialects. 
Characters have different backgrounds and appear in numerous environments or time periods. For this reason, we elected to 
train LaTextGAN on the Toronto Book Corpus, a collection of books known for both sentence quality and quantity \cite{moviebook}. We select two million sentences from the corpus for our training set.


We use the neural language model (NLM) and the variational autoencoder (VAE) as baselines. 


\subsection{Human Discriminators}

Evaluation of generative models remains a difficult endeavor. Traditionally, sentence quality has been evaluated with metrics such as METEOR and BLEU score \cite{papineni2002bleu}. However, these metrics often fail to evaluate the higher-level meaning of generated sentences. We propose an empirical evaluation using humans as discriminators, to differentiate between generated sentences and sentences from the dataset. 
To perform the evaluation, a pair of real and generated sentences is presented in random order to the participant. The most "realistic" sentence is selected from the pair, or both are selected as realistic. If both are selected as about equally realistic, this counts as a draw. See Table \ref{table:human_eval} for results. 

\subsection{BLEU Score}

For automatic evaluation, we calculate the BLEU-4 score of system-generated sentences against a validation set of 10,000 held-out sentences from the Toronto corpus. The variational autoencoder performed best on this metric, followed by LaTextGAN and the neural language model.

\begin{figure}[t!]
\begin{center}
\includegraphics[scale=.5]{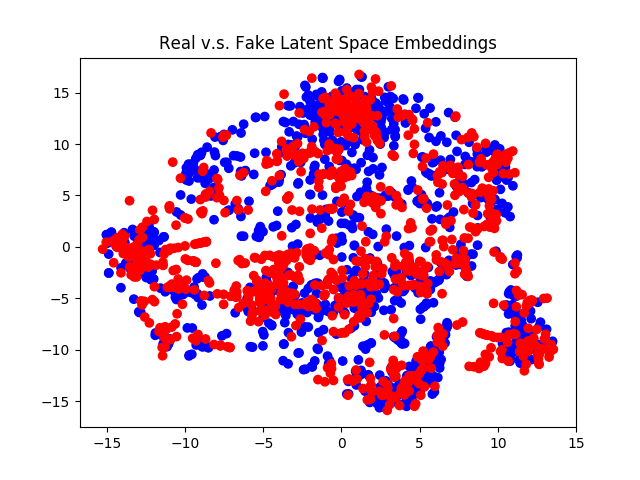}
\vspace{-2em}
\caption{Plot of GAN-generated sentence vectors (red) and genuine sentence vectors (blue) produced by the autoencoder. Dimensionality reduced using t-SNE.}
\label{fig:tsne}
\end{center}
\end{figure}

\subsection{Plotted Sentences}
A quality generator should produce sentence representations that lie in similar neighborhoods to real sentences within the autoencoder-learned latent space. To examine how our generator output compares, we plot both real and generated sentences using t-SNE \cite{maaten2008visualizing}. 
To examine the input space of our generator, we select two random points within the input z-space of the generator, and decode sentence vectors at evenly spaced intervals between them. Points that are close in the latent space should decode to similar sentences.

\section{Discussion}



The LaTextGAN system scored the highest percentage of sentences that were
rated as more realistic than a given English sentence from the dataset. However, the neural language model scored a higher draw rate. Judging by the dataset quality, it is possible that LaTextGAN is more likely to generate high quality sentences, while the neural language model is more likely to generate decent sentences that the evaluators deemed about as real as sentences from the dataset (medium quality, high probability).  While the VAE had the highest BLEU score, it was not competitive with the other models for human-rated realism. Our system performs well on both human and automatic evaluation. 

Unlike VAEs, the LaTextGAN input space is forced to be a Gaussian distribution, providing theoretical guarantees for the validity of high-probability regions of the space. We show sentences sampled from this space in Table \ref{table:examples}. Interpolated sentences reveal smooth transformations from one sentence to another. However, not all adjacent sentences are semantically similar (e.g. "They sat on captain" and "He looked at her") even if they share similar syntactic structure. Improving the GAN input space still remains one possible research direction. An encoder which maps generated sentence vectors back into Gaussian vectors could improve the smoothness and potential applications of the space \cite{ulyanov2017takes}.

The structure of LaTextGAN allows for more interesting evaluation techniques. Figure \ref{fig:tsne} contains sentence embeddings produced by our model, alongside genuine sentence embeddings encoded using the autoencoder. LaTextGAN-generated sentence embeddings cover every region of high-probability. This provides evidence that LaTextGAN does not suffer from mode-collapse as is often observed in GAN models.

\section{Conclusion}

We propose LaTextGAN as a new generative adversarial architecture for text, capable of being trained without reinforcement signals of any kind. We demonstrate that our GAN performs well on both human evaluation and BLEU-4 scoring, and achieves the highest percentage of sentences which rate higher quality than real sentences sampled from the dataset. Plotted embeddings indicate that LaTextGAN has properly modeled the latent space of a trained autoencoder, while sentence interpolations show that the LaTextGAN generator smoothly encodes sentences in the input space. 
%
We expect this model to be useful in downstream tasks such as dialogue generation where it would potentially increase response diversity.

In short, reinforcement learning poses a significant barrier-to-entry for the application of GAN models in natural language generation. We introduce a model which removes this barrier, with the aim of inspiring more widespread use of GANs outside of computer vision.



\bibliography{naaclhlt2018}

\begin{thebibliography}{25}
\expandafter\ifx\csname natexlab\endcsname\relax\def\natexlab#1{#1}\fi

\bibitem[{Arjovsky et~al.(2017)Arjovsky, Chintala, and
  Bottou}]{arjovsky2017wasserstein}
Martin Arjovsky, Soumith Chintala, and L{\'e}on Bottou. 2017.
\newblock Wasserstein generative adversarial networks.
\newblock In \emph{International Conference on Machine Learning}, pages
  214--223.

\bibitem[{Bengio et~al.(2001)Bengio, Ducharme, and Vincent}]{bengio2001neural}
Yoshua Bengio, R{\'e}jean Ducharme, and Pascal Vincent. 2001.
\newblock A neural probabilistic language model.
\newblock In \emph{Advances in Neural Information Processing Systems}, pages
  932--938.

\bibitem[{Bowman et~al.(2015)Bowman, Vilnis, Vinyals, Dai, Jozefowicz, and
  Bengio}]{bowman2015generating}
Samuel~R Bowman, Luke Vilnis, Oriol Vinyals, Andrew~M Dai, Rafal Jozefowicz,
  and Samy Bengio. 2015.
\newblock Generating sentences from a continuous space.
\newblock \emph{arXiv preprint arXiv:1511.06349}.

\bibitem[{Gatt and Krahmer(2018)}]{gatt2018survey}
Albert Gatt and Emiel Krahmer. 2018.
\newblock Survey of the state of the art in natural language generation: Core
  tasks, applications and evaluation.
\newblock \emph{Journal of Artificial Intelligence Research}, 61:65--170.

\bibitem[{Goodfellow et~al.(2014)Goodfellow, Pouget-Abadie, Mirza, Xu,
  Warde-Farley, Ozair, Courville, and Bengio}]{goodfellow2014generative}
Ian Goodfellow, Jean Pouget-Abadie, Mehdi Mirza, Bing Xu, David Warde-Farley,
  Sherjil Ozair, Aaron Courville, and Yoshua Bengio. 2014.
\newblock Generative adversarial nets.
\newblock In \emph{Advances in neural information processing systems}, pages
  2672--2680.

\bibitem[{Gulrajani et~al.(2017)Gulrajani, Ahmed, Arjovsky, Dumoulin, and
  Courville}]{gulrajani2017improved}
Ishaan Gulrajani, Faruk Ahmed, Martin Arjovsky, Vincent Dumoulin, and Aaron
  Courville. 2017.
\newblock Improved training of wasserstein gans.
\newblock \emph{arXiv preprint arXiv:1704.00028}.

\bibitem[{He et~al.(2016)He, Zhang, Ren, and Sun}]{he2016deep}
Kaiming He, Xiangyu Zhang, Shaoqing Ren, and Jian Sun. 2016.
\newblock Deep residual learning for image recognition.
\newblock In \emph{Proceedings of the IEEE conference on computer vision and
  pattern recognition}, pages 770--778.

\bibitem[{Hochreiter and Schmidhuber(1997)}]{hochreiter1997long}
Sepp Hochreiter and J{\"u}rgen Schmidhuber. 1997.
\newblock Long short-term memory.
\newblock \emph{Neural computation}, 9(8):1735--1780.

\bibitem[{Johnson et~al.(2016)Johnson, Schuster, Le, Krikun, Wu, Chen, Thorat,
  Vi{\'e}gas, Wattenberg, Corrado et~al.}]{johnson2016google}
Melvin Johnson, Mike Schuster, Quoc~V Le, Maxim Krikun, Yonghui Wu, Zhifeng
  Chen, Nikhil Thorat, Fernanda Vi{\'e}gas, Martin Wattenberg, Greg Corrado,
  et~al. 2016.
\newblock Google's multilingual neural machine translation system: enabling
  zero-shot translation.
\newblock \emph{arXiv preprint arXiv:1611.04558}.

\bibitem[{Kingma and Ba(2014)}]{kingma2014adam}
Diederik Kingma and Jimmy Ba. 2014.
\newblock Adam: A method for stochastic optimization.
\newblock \emph{arXiv preprint arXiv:1412.6980}.

\bibitem[{Lamb et~al.(2016)Lamb, GOYAL, Zhang, Zhang, Courville, and
  Bengio}]{lamb2016professor}
Alex~M Lamb, Anirudh Goyal ALIAS~PARTH GOYAL, Ying Zhang, Saizheng Zhang,
  Aaron~C Courville, and Yoshua Bengio. 2016.
\newblock Professor forcing: A new algorithm for training recurrent networks.
\newblock In \emph{Advances In Neural Information Processing Systems}, pages
  4601--4609.

\bibitem[{Ledig et~al.(2016)Ledig, Theis, Husz{\'a}r, Caballero, Cunningham,
  Acosta, Aitken, Tejani, Totz, Wang et~al.}]{ledig2016photo}
Christian Ledig, Lucas Theis, Ferenc Husz{\'a}r, Jose Caballero, Andrew
  Cunningham, Alejandro Acosta, Andrew Aitken, Alykhan Tejani, Johannes Totz,
  Zehan Wang, et~al. 2016.
\newblock Photo-realistic single image super-resolution using a generative
  adversarial network.
\newblock \emph{arXiv preprint}.

\bibitem[{Li et~al.(2017)Li, Monroe, Shi, Ritter, and
  Jurafsky}]{li2017adversarial}
Jiwei Li, Will Monroe, Tianlin Shi, Alan Ritter, and Dan Jurafsky. 2017.
\newblock Adversarial learning for neural dialogue generation.
\newblock \emph{arXiv preprint arXiv:1701.06547}.

\bibitem[{Maaten and Hinton(2008)}]{maaten2008visualizing}
Laurens van~der Maaten and Geoffrey Hinton. 2008.
\newblock Visualizing data using t-sne.
\newblock \emph{Journal of machine learning research}, 9(Nov):2579--2605.

\bibitem[{Makhzani et~al.(2015)Makhzani, Shlens, Jaitly, Goodfellow, and
  Frey}]{makhzani2015adversarial}
Alireza Makhzani, Jonathon Shlens, Navdeep Jaitly, Ian Goodfellow, and Brendan
  Frey. 2015.
\newblock Adversarial autoencoders.
\newblock \emph{arXiv preprint arXiv:1511.05644}.

\bibitem[{Papineni et~al.(2002)Papineni, Roukos, Ward, and
  Zhu}]{papineni2002bleu}
Kishore Papineni, Salim Roukos, Todd Ward, and Wei-Jing Zhu. 2002.
\newblock Bleu: a method for automatic evaluation of machine translation.
\newblock In \emph{Proceedings of the 40th annual meeting on association for
  computational linguistics}, pages 311--318. Association for Computational
  Linguistics.

\bibitem[{Rush et~al.(2015)Rush, Chopra, and Weston}]{rush2015neural}
Alexander~M Rush, Sumit Chopra, and Jason Weston. 2015.
\newblock A neural attention model for abstractive sentence summarization.
\newblock \emph{arXiv preprint arXiv:1509.00685}.

\bibitem[{Serban et~al.(2016)Serban, Sordoni, Bengio, Courville, and
  Pineau}]{serban2016building}
Iulian~Vlad Serban, Alessandro Sordoni, Yoshua Bengio, Aaron~C Courville, and
  Joelle Pineau. 2016.
\newblock Building end-to-end dialogue systems using generative hierarchical
  neural network models.
\newblock In \emph{AAAI}, volume~16, pages 3776--3784.

\bibitem[{Serban et~al.(2017)Serban, Sordoni, Lowe, Charlin, Pineau, Courville,
  and Bengio}]{serban2017hierarchical}
Iulian~Vlad Serban, Alessandro Sordoni, Ryan Lowe, Laurent Charlin, Joelle
  Pineau, Aaron~C Courville, and Yoshua Bengio. 2017.
\newblock A hierarchical latent variable encoder-decoder model for generating
  dialogues.
\newblock In \emph{AAAI}, pages 3295--3301.

\bibitem[{Ulyanov et~al.(2017)Ulyanov, Vedaldi, and
  Lempitsky}]{ulyanov2017takes}
Dmitry Ulyanov, Andrea Vedaldi, and Victor Lempitsky. 2017.
\newblock It takes (only) two: Adversarial generator-encoder networks.
\newblock \emph{arXiv preprint arXiv:1704.02304}.

\bibitem[{Weissenborn et~al.(2017)Weissenborn, Wiese, and
  Seiffe}]{weissenborn2017making}
Dirk Weissenborn, Georg Wiese, and Laura Seiffe. 2017.
\newblock Making neural qa as simple as possible but not simpler.
\newblock \emph{arXiv preprint arXiv:1703.04816}.

\bibitem[{Yu et~al.(2017)Yu, Zhang, Wang, and Yu}]{yu2017seqgan}
Lantao Yu, Weinan Zhang, Jun Wang, and Yong Yu. 2017.
\newblock Seqgan: Sequence generative adversarial nets with policy gradient.

\bibitem[{Zhang et~al.(2017{\natexlab{a}})Zhang, Xu, Li, Zhang, Huang, Wang,
  and Metaxas}]{zhang2017stackgan}
Han Zhang, Tao Xu, Hongsheng Li, Shaoting Zhang, Xiaolei Huang, Xiaogang Wang,
  and Dimitris Metaxas. 2017{\natexlab{a}}.
\newblock Stackgan: Text to photo-realistic image synthesis with stacked
  generative adversarial networks.
\newblock In \emph{IEEE Int. Conf. Comput. Vision (ICCV)}, pages 5907--5915.

\bibitem[{Zhang et~al.(2017{\natexlab{b}})Zhang, Gan, Fan, Chen, Henao, Shen,
  and Carin}]{zhang2017adversarial}
Yizhe Zhang, Zhe Gan, Kai Fan, Zhi Chen, Ricardo Henao, Dinghan Shen, and
  Lawrence Carin. 2017{\natexlab{b}}.
\newblock Adversarial feature matching for text generation.
\newblock \emph{arXiv preprint arXiv:1706.03850}.

\bibitem[{Zhu et~al.(2015)Zhu, Kiros, Zemel, Salakhutdinov, Urtasun, Torralba,
  and Fidler}]{moviebook}
Yukun Zhu, Ryan Kiros, Richard Zemel, Ruslan Salakhutdinov, Raquel Urtasun,
  Antonio Torralba, and Sanja Fidler. 2015.
\newblock Aligning books and movies: Towards story-like visual explanations by
  watching movies and reading books.
\newblock In \emph{arXiv preprint arXiv:1506.06724}.

\end{thebibliography}
\bibliographystyle{acl_natbib_nourl}

\appendix

\section{Model Implementation}
\label{sec:supplemental}

Here we report model parameters and implementation decisions non-central to the paper. It is well-documented that GANs suffer from a number of convergence problems \cite{goodfellow2014generative}. To solve these issues, \newcite{arjovsky2017wasserstein} propose the Wasserstein GAN (WGAN), which uses the Earth-Mover distance metric for optimization, and clips the discriminator weights during training. They show the Earth-Mover distance provides stable gradients for all points in the solution space. \newcite{gulrajani2017improved} introduce a regularization of the WGAN which encourages the norm of discriminator gradients to approach unity. They show the regularization increases the capacity of larger models, including ResNets. We choose to implement a WGAN model with this regularization to avoid convergence issues associated with the vanilla GAN architecture.

Empirically, we choose the cell sizes for the encoder and decoder LSTMs of our autoencoder to be 100 and 600, respectively. We apply a dropout of 0.5 to the encoder output during training. The application of dropout improved the quality of generated sentences. Input words to the encoder are represented by 200-dimensional word embeddings, learned during training.

ResNet architectures can be deeper than standard fully-connected layers. We implement the generator
and discriminator ResNets using 40 layers each. Each layer is of the form $F(x) = H(x) + x$ where $H(x)$ is the learned residual layer function. Note that for a learned fully-connected function $G(x)$, a residual layer can learn $H(x)=G(x) - x$, giving residual layers an equivalent capacity to standard dense layers. In both generator and discriminator layers, we choose to implement each residual layer as

\begin{equation}
H(x)=relu(x \cdot W_1 + b_1) \cdot W_2 + b_2
\end{equation}

\noindent
for learned weight matrices $W_1$, $W_2$ and biases $b_1$, $b_2$. For simplicity, all layers have the same dimension of 100.

We implement our NLM as an LSTM decoder network with the same size (600-dimensional cell state) and configuration as the decoder of our autoencoder network, for fair comparison. We sample words from their probabilities at each time-step to decode each full sequence. For our variational autoencoder, we select the encoder and decoder LSTM networks to be 600-dimensional, and apply KL annealing to enhance training \cite{bowman2015generating}. 200-dimensional word embeddings are learned as input. We use learning rates of $10^{-3}$ and $5\cdot10^{-4}$ for the NLM and VAE respectively and train for 5 epochs.

As is suggested in the paper, we update the discriminator N times per update to the generator. In practice, we cannot train a fully optimal discriminator, so we use N=10. Adam optimizer is used for optimization of all models for its fast convergence properties \cite{kingma2014adam}. We use a learning rate $\alpha=5 \cdot 10^{-4}$ for the autoencoder and $\alpha=1 \cdot 10^{-4}$ for both the generator and discriminator. We train the autoencoder and WGAN for 5 and 15 epochs, respectively. For the stochastic input to the generator $p(z)$, we sample vectors from a multivariate Gaussian distribution.

To improve the quality of our data, we filter vocabulary words which appear less than 5 times and remove all sentences which contain these rare words. Sentences are restricted to a maximum length of 20 words. 

For human evaluation, we assembled 1,000 sentence pairs for human evaluation (one third per model), and selected two unaffiliated evaluators to label sentence pairs. Each pair was labeled once. The order of pairs were randomized, and the real and generated sentence within each pair was randomized. All pairs were completed in a few hours, highlighting this method as a feasible human evaluation for future generative models. BLEU-4 score was calculated using a held-out validation set of 10,000 sentences, and was calculated using weights (0.25, 0.25, 0.25, 0.25) for 1, 2, 3, and 4-grams respectively. BLEU scores were calculated for 1000 sentences generated from each model, and averaged.

To create sentence interpolations, two Gaussian vectors $v_1$ and $v_2$ were sampled as starting points in the input space. We then sample N-1 points linearly between them, each point calculated as

\begin{equation}
v_i=v_1 + (v_2-v_1)/N * i
\end{equation}

\noindent
for all intermediate points $i=1, 2, 3 ... N-1$. Each of these points is used as input to the generator and decoded to produce a sentence.

\end{document}